\documentclass[letterpaper, 12 pt, conference]{ieeeconf}  % Comment this line out
\IEEEoverridecommandlockouts                              % This command is only
\usepackage{graphicx} % for pdf, bitmapped graphics files
\usepackage{bm}

\usepackage{etoolbox}
\usepackage{float}
\usepackage{caption}
\usepackage{amsmath}
\usepackage{array, multirow}
\usepackage{adjustbox}
\usepackage{flushend}
\newcommand{\RN}[1]{%
  \textup{\lowercase\expandafter{\romannumeral#1}}%
}
\renewcommand{\theenumi}{\roman{enumi}}
\patchcmd{\abstract}{Abstract}{Purpose}{}{}

\title{\LARGE \bf
Occupancy Detection in Room Using Sensor Data}

\author{Mohammadhossein Toutiaee\\
hossein@uga.edu}

\begin{document}

\maketitle
\thispagestyle{plain}
\pagestyle{plain}

%%%%%%%%%%%%%%%%%%%%%%%%%%%%%%%%%%%%%%%%%%%%%%%%%%%%%%%%%%%%%%%%%%%%%%%%%%%%%%%%
\begin{abstract}
With the advent of Internet of Thing (IoT), and ubiquitous data collected every moment by either portable (smart phone) or fixed (sensor) devices, it is important to gain insights and meaningful information from the sensor data in context-aware computing environments. Many researches have been implemented by scientists in different fields, to analyze such data for the purpose of security, energy efficiency, building reliability and smart environments. One study, that many researchers are interested in, is to utilize Machine Learning techniques for occupancy detection where the aforementioned sensors gather information about the environment. This paper provides a solution to detect occupancy using sensor data by using and testing several variables. Additionally we show the analysis performed over the gathered data using Machine Learning and pattern recognition mechanisms is possible to determine the occupancy of indoor environments. Seven famous algorithms in Machine Learning, namely as Decision Tree, Random Forest, Gradient Boosting Machine, Logistic Regression, Naive Bayes, Kernelized SVM and K-Nearest Neighbors are tested and compared in this study.
\end{abstract}

%%%%%%%%%%%%%%%%%%%%%%%%%%%%%%%%%%%%%%%%%%%%%%%%%%%%%%%%%%%%%%%%%%%%%%%%%%%%%%%%
\section{Introduction}
Sensor data are everywhere, and the amount of data generated from such ubiquitous sensors in the heterogeneous distributed systems has increased significantly. One aspect of such data is to predict the environment and the human behaviours. With great success of Machine Learning applications in many domains, people are more interested in applying Artificial Intelligence techniques and Statistical Learning methods into the performance analysis, to predict and evaluate their systems. Additionally, one can contribute towards sustainable energy by incorporating better energy consumption practices especially in new construction. Or one can specify whether or not a particular space, room or building being occupied by people, and a following question could be the number of people in the target place. Hence, occupancy can be detected by analyzing the sensor data. Another example of such systems is the Management of Energy Consumption and Heat Ventilation and Air Conditioning (HVAC) \cite{jin2017virtual,imanishi2015enhanced} in Smart Buildings. To be specific, information from occupancy detection and occupant's behavioral patterns can be used to manage and control buildings more intelligently for ventilation, heating and cooling, and energy efficiency \cite{ryu2016development,leephakpreeda2005adaptive}.
The dataset used for this research has been collected by \cite{candanedo2016accurate}, and has used data recorded from light, temperature, humidity and CO2 sensors as variables to alarm occupancy and one camera to label ground truth for supervised classification model training. One can possibly include or combine those variables in the model to maximize the prediction accuracy. The Machine Learning models for this study are limited to Decision Tree, Random Forest, Gradient Boosting Machine, Logistic Regression, Naive Bayes, Kernelized SVM and K-Nearest Neighbors, and all the trained and tested models are evaluated on these aforementioned algorithms. These statistical methods are already being used in many open source platforms such as Python and R. This research will implement the models in Python, and the use and performance of all those proposed models have been reported previously in other scientific literature for occupancy detection task.
\begin{figure}[H]
\includegraphics[width=\linewidth]{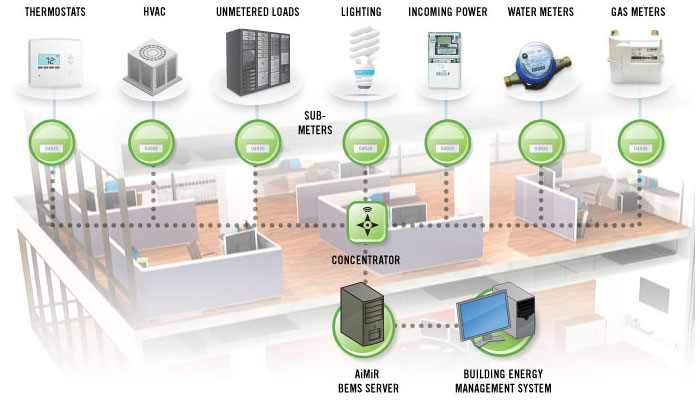}
\caption*{\small{credit: www.nuritelecom.com}}\label{exp}
\end{figure}
\section{Related Work}
With the great influence of IoT, many technologies are built over the pervasive systems from a variety of computing environments. Some technologies would take the advantage of accessibility sensors such as wearable systems. These systems are known to be cheap and omnipresent, and people are comfortable to use them. Smart phones and smart watches are examples in this regard. On the other hand, we believe that fixed and targeted sensors are more popular among people who are interested in smart environments. Occupancy detection systems are classified based on the dependency of the system to a terminal. In \cite{lee2006pyroelectric} and \cite{labeodan2015occupancy}\nocite{toutiaee2101gaussian,toutiaee2020stereotype,peng2019knowledge,toutiaee2020video} it is mentioned that in the cases that require a terminal, it is necessary to attach a device to the occupants to keep track of them (e.g., a smart phone). In the non-terminal methods, the detection is based on a passive approach that is focused on monitoring areas or spaces instead of the identification of devices. In both settings, data are collected in stream fashion and are buffered in a middle layer interface to which the interface is connected. Although most part of the implementation is depending on the data gathering from the sensors, all prediction models can be applied either online or offline, based on the requirements of the system demand. In the online mode, the models are trained using streaming data, with some tolerable latency, while the offline model will be implemented in such a way that the model is pushed to train later once the amount of data is available. Both modes have advantages and disadvantages, however, this paper would not address them.\\
Different measurement techniques are used for occupation detection in various studies. The occupation detection techniques can be classified as:
\begin{enumerate}
\item Occupancy measurement via camera
\item Occupancy measurement via passive infra-red (PIR) sensor
\item Occupancy via ultrasonic sensor
\item Occupancy measurement via radio frequency (RF) signals
\item Occupancy measurement via sensors fusion
\item Occupancy measurement via WLAN, Bluetooth and WiFi
\end{enumerate}
This study has used a dataset containing light, temperature, humidity and CO2 variables as a proxy to detect occupancy and a camera to establish ground occupancy for a supervised classification model. Some of the Machine Learning methods used in this paper are also utilized in \cite{candanedo2016accurate}. This article is not implementing a new approach for occupancy detection problem, but rather trying to reproduce the work that \cite{candanedo2016accurate} has already performed in their article as well as other machine leaning approaches not in \cite{candanedo2016accurate}. We aim to compare how similar (better/worse) one can obtain out of a same dataset, with the same method in \cite{candanedo2016accurate} and additional ones.
\section{Summary of Data}
The dataset provided for this study contains three separate files; training, validation and testing. The validation dataset was collected in a open-door situation. Also, the test set was gathered in a closed-door situation. The training, validation and test sets consist of 8143, 2665 and 9752, respectively. All the sets contain seven variables; Temperature, Humidity, Light, CO2, Humidity, Ratio and Occupancy, where the Occupancy is the response variable indicated with 0 and 1 (occupied/empty). The summary statistics for the three aforementioned data-sets are in the Appendix (Figures \ref{tr}, \ref{val} and \ref{ts}). One can observe from the tables that the maximum value for CO2 in the Training and Test sets are similar (2028 and 2076), whereas one can see a lower value in the Validation set. This is because the Validation set was collected in an open-door environment and air could ventilate in the room freely and the CO2 level has been affected accordingly. This fact is also true for the Humidity variable as the Validation set shows a relatively lower value than the Training and Test sets. Moreover, one can see that the Light variable has been affected heavily when the door is open or closed. This amount of variation in Light could help classify the occupancy of the room better than other variables, since people will more likely keep the door opened while they are in the room. The rest of the information is visible in the Appendix as mentioned.
\section{Analysis}
\subsection{Exploratory Analysis}
Figure \ref{fig1} in Appendix shows how variables are correlated to one another. As one can observe, the ``Light'' variable is a good predictor for this dataset, since there is a distinct separation between two classes (1 and 0). Also the plot shows that the Humidity Ratio and the Humidity are highly correlated, and CO2 and Humidity together are useless since both pairs are highly correlated. Moreover, the Temperature- CO2, Temperature-Humidity and Temperature-Humidity Ratio are not contributing to the response variable at all. The plot indicates that the ``Light'' with almost all of the variables can classify the model quite well. 
\subsubsection{Occupancy Variable}
One can also see the time series per predictor (Figure \ref{fig2}). As it is visible in the time series plots, one would observe a large gap between two dates (07-02-2015 and 09-02-2015). This is due to the fact that those two days are  weekend, so it is expected that nobody shows up in the office during those days. Although this is a very strong predictor, this study does not use it in the models, so that the effects of other sensor variables become noted.
\subsubsection{Light Variable}
The Light variable appears to be less than 400lx in the weekend (Figure 3). Light follows the same pattern as the Occupancy. Conspicuously enough, there is an abrupt signal in the lighting in the weekend, which may be outliers.
\subsubsection{CO2 Variable}
This very variable looks similar to both Occupancy and Light variables. One can observe an oscillation when there is an occupant in the room. 
\section{Model Setup and Implementation}
Having analyzed different variables in the dataset, one can propose several popular models, so that one would see the performance of the models in this study. This is a very conventional setup in Machine Learning field in which one would split the dataset into the Training, Validation and Test sets. The Training set contains all the aforementioned variables except Occupancy, where this is the response variable. This is applicable in the Validation and Test sets where one would include all the variables except the response variable. Additionally, one would exclude the "Date" variable from the model, as it was discussed in the previous section, since that is not a sensor data variable, however, in other studies they have trained their models including the Date variable.\\
\indent A list of tuples for predictors are declared in the Training models, so that one can observe them in testing one's models with different feature combinations. The proposed techniques for training the models are as follow; Decision Tree (Table \ref{tab:DT}), Random Forest (Table \ref{tab:RF}), Gradient Boosting Machine (Table \ref{tab:GB}), Logistic Regression (Table \ref{tab:LR}), Naive Bayes (Table \ref{tab:NB}), Kernelized SVM (Table \ref{tab:SVM}) and K-Nearest Neighbors (Table \ref{tab:KN}). The detailed information about how well each algorithm performs over the Validation and Test sets are provided in the Appendix. Among those methods, one can see that the predictors perform reasonably well in L2 (0.99) but not very promising in L1 penalties (0.81). In Naive Bayes, one would see that Light and CO2 together might predict the Occupancy with \%98 of accuracy, which is very satisfying, while CO2 with Temperature do not perform well in the model. One might be interested in knowing the accuracy for the K-NN approach with similar variables (Light and CO2) in the Validation and Test sets, which are \%98 and \%97 of accuracy, respectively. The Decision Tree model produced more than \%98 accuracy for the triplet Training, Validation and Test sets, which is ranked the best up to this point. The RF model surpassed the DT model also, with similar accuracies to DT model, in that both models indicate a high accuracy with Light and CO2, and RF has a higher accuracy than DT with CO2 and Temperature variables. One would see the accuracy in GBM with \%99 in the Training and Test sets, while not performing well in the CO2 and Temperature combinations. And last but not least, the K-SVM performs interestingly similar to RF model in that both show a high accuracy for Light-CO2 and CO2-Temperature combinations. 
\section{Conclusion}
Researchers and industries are trying to improve the life quality of people, using the IoT paradigm and pervasive systems to reach the idea of building Smart Environments. One aspect of ubiquitous computing is to apply Artificial Intelligence techniques in the aforementioned environments. Machine Learning, as a core methodology, will be applicable in many domains including citizen science. This study has shown that it is possible to obtain high accuracies in the determination of occupancy with several statistical learning models. Additionally, most classifiers have shown satisfying results using Light-CO2 alone (Table (\ref{tab:conc1})). One can extend this study to consider other variables, or to construct new variables from the existing ones. Some models had advantages and disadvantages in applying them in the micro-controllers or cheap processors. 
\subsection{Future Work}
One interesting study is to find the number of occupants in the room or building. There are several articles around this field, each of which uses different techniques. Another interesting research could be incorporating smart phone's sensors in the training models, and fuse them with the current data-set. To the best of our knowledge, not many articles have been found around this domain, and this problem (and similar ones), could be extended to the current state-of-the-art techniques.
\bibliographystyle{IEEEtran}
\bibliography{IEEEabrv,main}
\newpage
\section*{Appendix}
\begin{figure}[H]
\includegraphics[width=\linewidth]{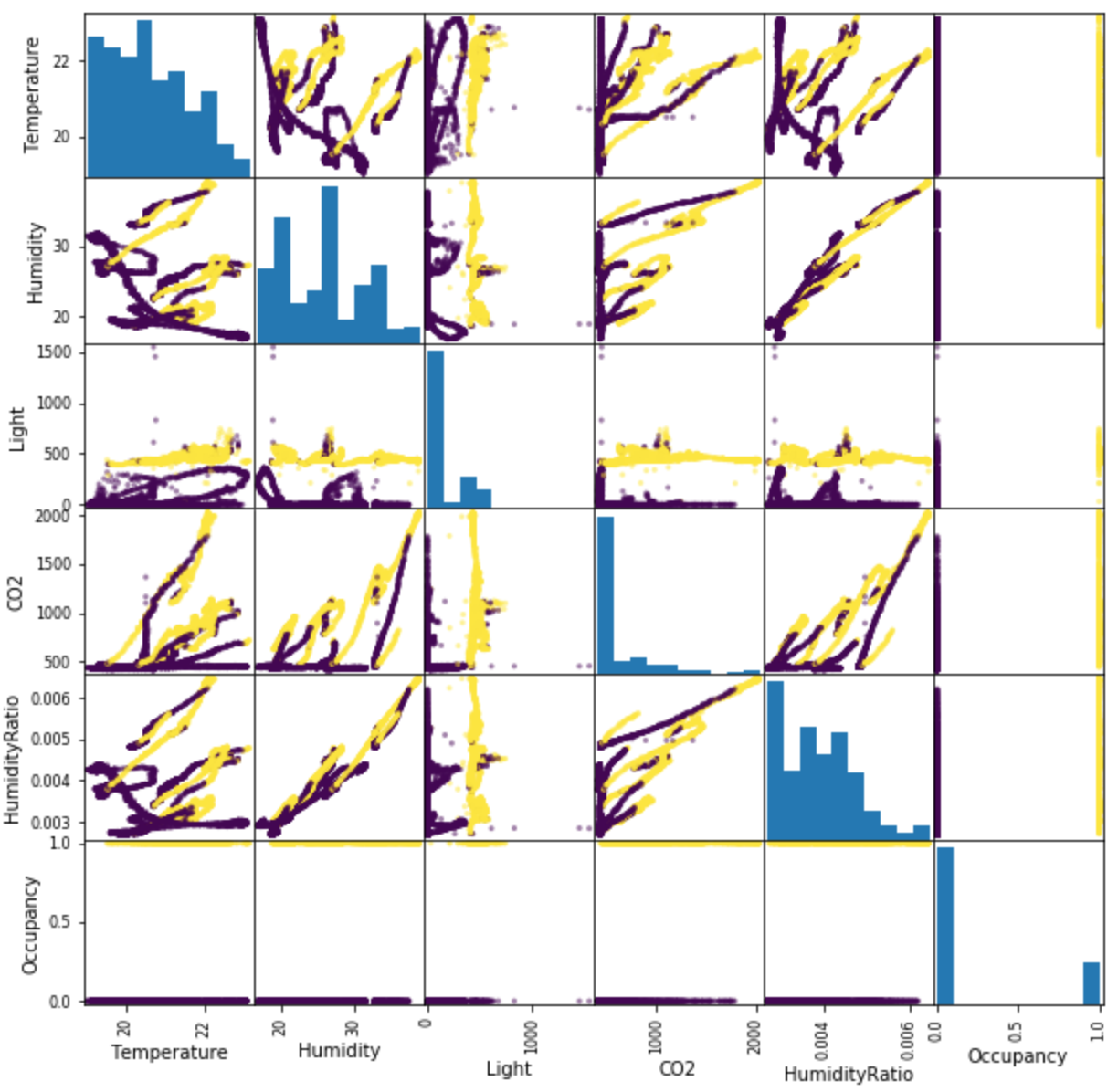}
\caption{\small{Variables correlations.}}\label{fig1}
\end{figure}
\begin{figure}[H]
\includegraphics[width=\linewidth]{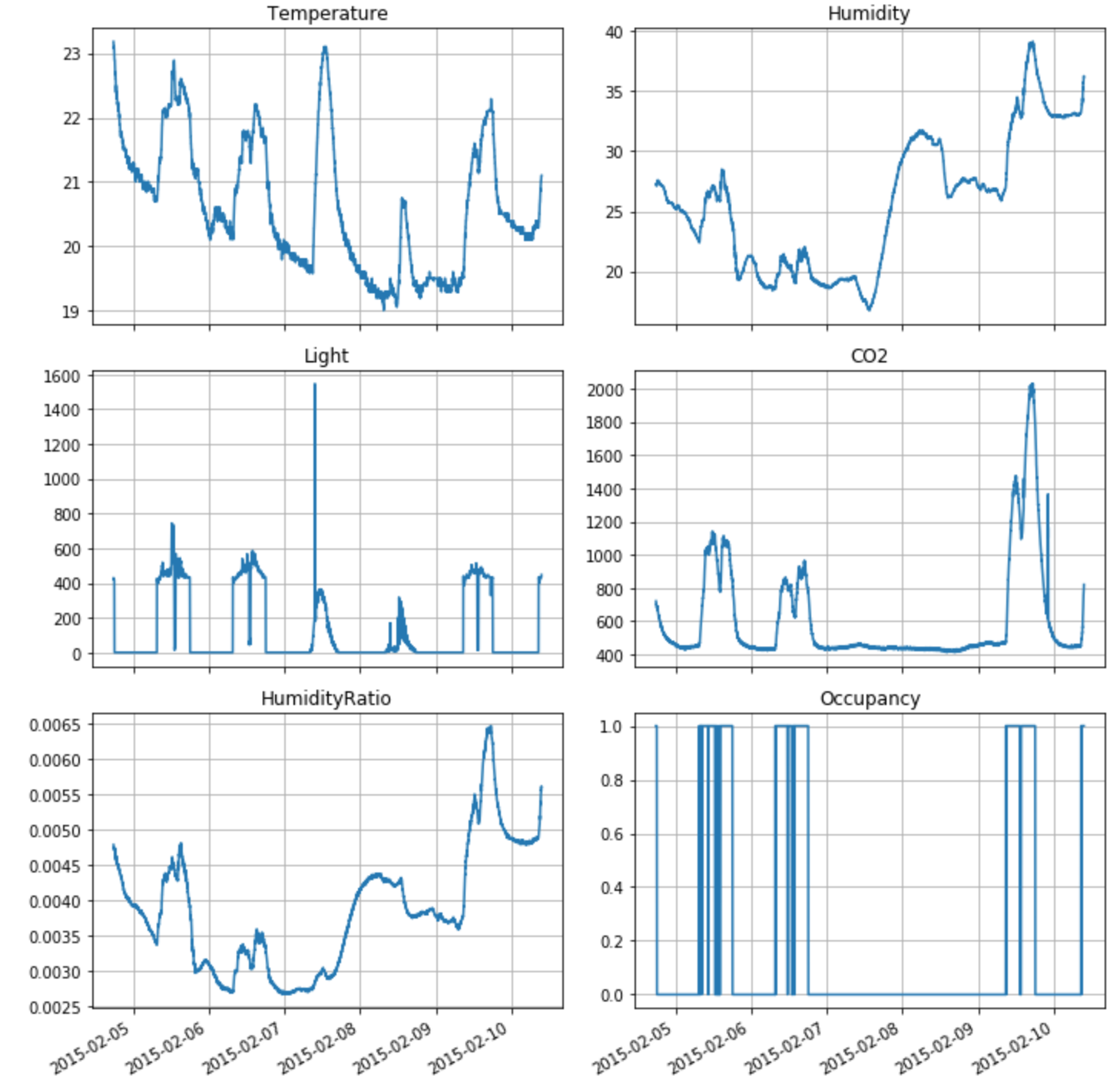}
\caption{\small{Time series per feature.}}\label{fig2}
\end{figure}
\newpage
\begin{table*}
\centering
\captionof{table}{Logistic Regression} \label{tab:LR} 
\begin{tabular}{lcccc}
  \hline
Features & Hyper Parameters & Train & Validation & Test \\ 
  \hline
Light - CO2 & \{'random\_state': 0, 'C': 1, 'penalty': 'l2'\} & 0.99 & 0.98 & 0.99 \\ 
CO2 - Temperature & \{'random\_state': 0, 'C': 1.5, 'penalty': 'l1'\} & 0.90 & 0.88 & 0.81 \\ 
   \hline \\
\end{tabular}
\end{table*}
\begin{table*}
\centering
\captionof{table}{Naive Bayes} \label{tab:NB} 
\begin{tabular}{lccc}
  \hline
Features  & Train & Validation & Test \\ 
  \hline
Light - CO2 & 0.9835 & 0.9925 & 0.9882 \\ 
CO2 - Temperature & 0.9183 & 0.9557 & 0.7671 \\ 
   \hline \\
\end{tabular}
\end{table*}
\begin{table*}
\centering
\captionof{table}{K-Nearest Neighbors} \label{tab:KN} 
\begin{tabular}{lcccc}
  \hline
Features  & Neighbors & Train & Validation & Test \\ 
  \hline
Light - CO2 & 33 & 0.99 & 0.98 & 0.97 \\ 
CO2 - Temperature & 49 & 0.93 & 0.86 & 0.79 \\ 
   \hline \\
\end{tabular}
\end{table*}
\begin{table*}
\centering
\captionof{table}{Decision Tree} \label{tab:DT} 
\begin{tabular}{lcccc}
  \hline
Features & Hyper Parameters & Train & Validation & Test \\ 
  \hline
Light - CO2 & \{'min\_samples\_split': 2, 'max\_depth': 1, 'random\_state': 0\} & 0.99 & 0.98 & 0.99 \\ 
CO2 - Temperature & \{'min\_samples\_split': 2, 'max\_depth': 1, 'random\_state': 0\} & 0.93 & 0.87 & 0.78 \\ 
   \hline \\
\end{tabular}
\end{table*}
\begin{table*}
\centering
\captionof{table}{Random Forest} \label{tab:RF} 
\begin{tabular}{lcccc}
  \hline
Features & Hyper Parameters & Train & Validation & Test \\ 
  \hline
Light - CO2 & \{'min\_samples\_split': 2, 'max\_depth': 2, 'n\_estimators': 18, 'random\_state': 0\} & 0.99 & 0.98 & 0.99 \\ 
CO2 - Temperature & \{'min\_samples\_split': 2, 'max\_depth': 1, 'n\_estimators': 15, 'random\_state': 0\} & 0.92 & 0.85 & 0.86 \\ 
   \hline \\
\end{tabular}
\end{table*}
\begin{table*}
\centering
\captionof{table}{Gradient Boosting Machine} \label{tab:GB} 
\begin{tabular}{lcccc}
  \hline
Features & Hyper Parameters & Train & Validation & Test \\ 
  \hline
Light - CO2 & \{'random\_state': 0, 'n\_estimators': 112, 'learning\_rate': 0.08\} & 0.99 & 0.94 & 0.99 \\ 
CO2 - Temperature & \{'random\_state': 0, 'n\_estimators': 111, 'learning\_rate': 0.01\} & 0.93 & 0.69 & 0.77 \\ 
   \hline \\
\end{tabular}
\end{table*}
\begin{table*}
\centering
\captionof{table}{Kernalized SVM} \label{tab:SVM} 
\begin{tabular}{lcccc}
  \hline
Features & Hyper Parameters & Train & Validation & Test \\ 
  \hline
Light - CO2 & \{'random\_state': 0, 'kernel': 'linear'\} & 0.99 & 0.98 & 0.99 \\ 
CO2 - Temperature & \{'random\_state': 0, 'kernel': 'linear'\} & 0.92 & 0.86 & 0.84 \\ 
   \hline \\
\end{tabular}
\end{table*}
\begin{table*}
\centering
\captionof{table}{Accuracy Table for Models using Light-CO2} \label{tab:conc1} 
\begin{tabular}{lccccccc}
  \hline
 Light-CO2 & LR & NB & K-NN & DT & RF & GBM & K-SVM \\ 
  \hline
Train Accuracy & 0.99 & 0.98 & 0.99 & 0.99 & 0.99 & 0.99 & 0.99 \\ 
Valid Accuracy & 0.98 & 0.99 & 0.98 & 0.98 & 0.98 & 0.94 & 0.98 \\
Test Accuracy & 0.99 & 0.98 & 0.97 & 0.99 & 0.99 & 0.99 & 0.99 \\
   \hline \\
\end{tabular}
\end{table*}
\newpage
\begin{figure*}
\centering
\includegraphics[width=0.89\linewidth]{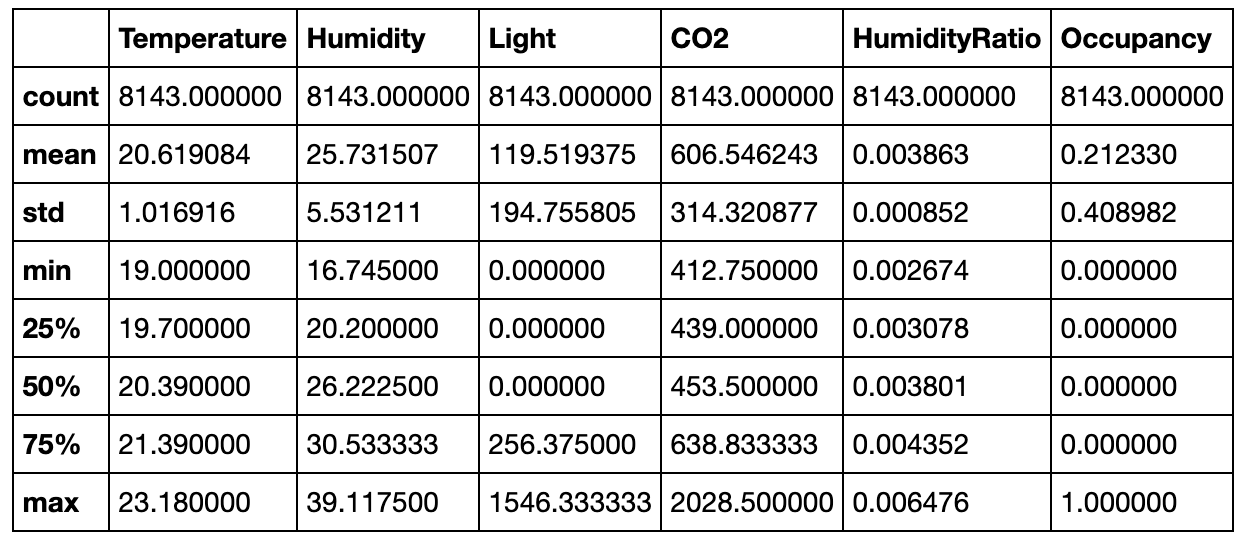}
\caption{\small{Summary Statistics - Training Set}}\label{tr}
\end{figure*}
\begin{figure*}
\centering
\includegraphics[width=0.89\linewidth]{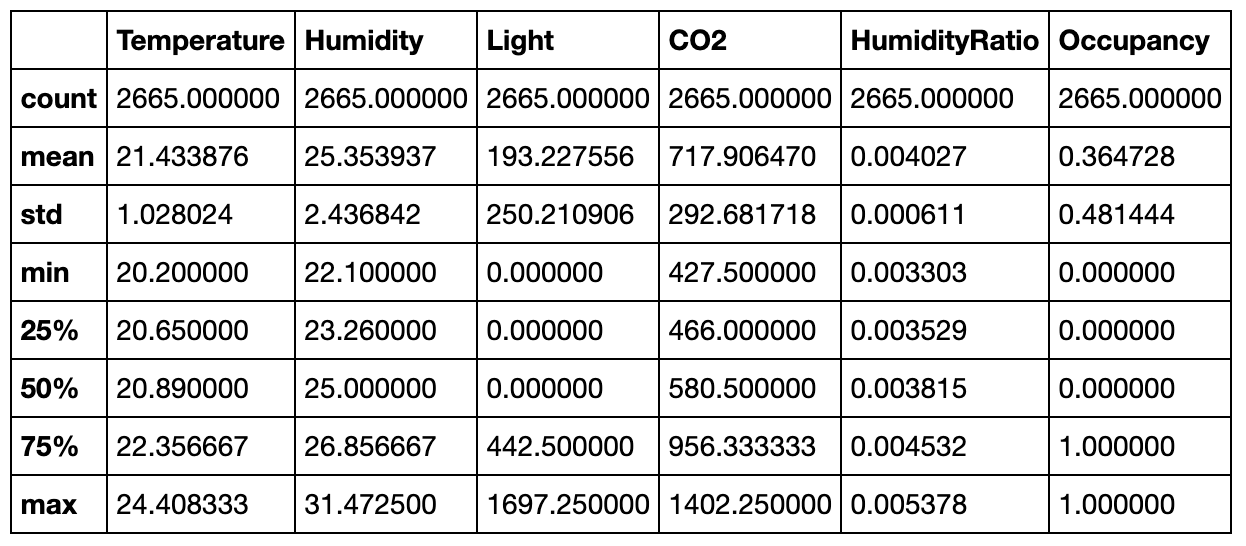}
\caption{\small{Summary Statistics - Validation Set}}\label{val}
\end{figure*}
\begin{figure*}
\centering
\includegraphics[width=0.89\linewidth]{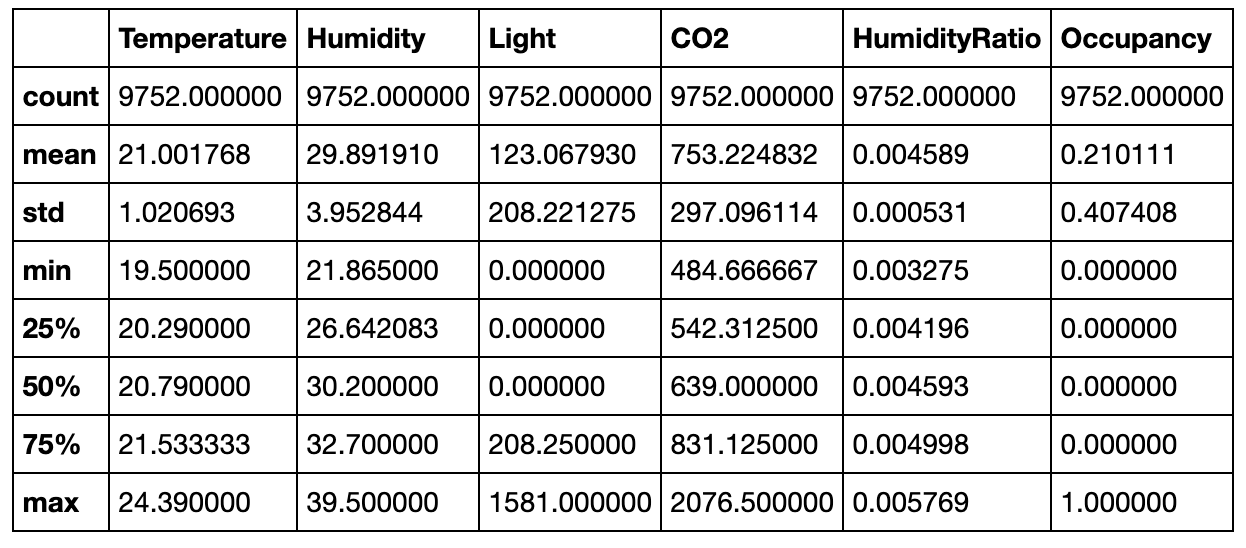}
\caption{\small{Summary Statistics - Test Set}}\label{ts}
\end{figure*}
\end{document}